\documentclass[conference,a4paper]{APSIPA2012}
\usepackage{multirow}
\usepackage{subcaption}
\usepackage{graphicx}
\usepackage[tbtags]{amsmath}

\usepackage[psamsfonts]{amssymb}
\usepackage{amsxtra}

\usepackage{bm}

\usepackage{cite}
\usepackage[utf8]{inputenc}

\graphicspath{ {imgs/} }

\DeclareMathOperator*{\argmax}{argmax}

\def\vec#1{\ensuremath{\bm{{#1}}}}
\def\mat#1{\vec{#1}}

\newcommand {\I} {\mat{I}}
\newcommand {\0} {\vec{0}}
\newcommand {\N} {\mathcal{N}}

\newcommand {\xn} {\vec{x}_n}
\newcommand {\yn} {\vec{y}_n}
\newcommand {\zn} {\vec{z}_n}

\newcommand {\zhat} {\vec{\hat{z}}_n}
\newcommand {\xhat} {\vec{\hat{x}}_n}

\newcommand {\thetas} {\vec{\theta}}
\newcommand {\phis} {\vec{\phi}}

\newcommand {\xsn} {\vec{x}_{s, n}}
\newcommand {\xtn} {\vec{x}_{t, m}}
\newcommand {\xsnhat} {\vec{\hat{x}}_{s, n}}
\newcommand {\xtnhat} {\vec{\hat{x}}_{t, m}}
\newcommand {\xnhat} {\vec{\hat{x}}_{n}}

\newcommand {\xsigma} {diag(\vec{\sigma}_{\xn})}

\newcommand {\eq}[1] {(\ref{#1})}

\newcommand {\eps} {\vec{\epsilon}_{n}}

\newcommand {\Xs} {\mat{X}_{s}}
\newcommand {\Xt} {\mat{X}_{t}}

\newcommand {\X} {\mat{X}}

\newcommand {\f} {f(\cdot)}
\newcommand {\p} {p_{\theta}}
\newcommand {\q} {q_{\phi}}

\newcommand {\pTrue} {p}

\newcommand {\ft} {f_{\theta}}
\newcommand {\fp} {f_{\phi}}

\newcommand {\phiOne} {\vec{\phi}_1}
\newcommand {\phiTwo} {\vec{\phi}_2}

\newcommand {\thetaOne} {\vec{\theta}_1}
\newcommand {\thetaTwo} {\vec{\theta}_2}

\newcommand {\ftOne} {f_{\thetaOne}}
\newcommand {\ftTwo} {f_{\thetaTwo}}
\newcommand {\fpOne} {f_{\phiOne}}
\newcommand {\fpTwo} {f_{\phiTwo}}

\newcommand {\half} {\dfrac{1}{2}}
\newcommand {\zvard} {\sigma^2_{\vec{z}_{n, d}} }
\newcommand {\zmud} {\mu_{\vec{z}_{n, d}}}

\newcommand {\zmu} {\vec{\mu}_{\zn}}
\newcommand {\zsig} {\vec{\sigma}_{\zn}}

\newcommand {\xvard} {\sigma^2_{\vec{x}_{n, d}}}
\newcommand {\xmud} {\mu_{\vec{x}_{n, d}}}

\newcommand {\xmu} {\vec{\mu}_{\xn}}
\newcommand {\xsig} {\vec{\sigma}_{\xn}}

\newcommand {\summation} {\sum\limits}

\newcommand {\KLD}[2] {D_{KL}(#1||#2)}
\newcommand {\E}[2] {\mathbf{E}_{#1}[#2]}
\newcommand {\LB}[1] {\mathcal{L}(\thetas, \phis; #1)}
\newcommand {\LBhat}[1] {\hat{\mathcal{L}}(\thetas, \phis; #1)}

\begin{document}

\title{Voice Conversion from Non-parallel Corpora Using Variational Auto-encoder}

\author{%
  \authorblockN{
    Chin-Cheng Hsu\authorrefmark{1}, 
    Hsin-Te Hwang\authorrefmark{1}, 
    Yi-Chiao Wu\authorrefmark{1}, 
    Yu Tsao\authorrefmark{2} and
    Hsin-Min Wang\authorrefmark{1}
  }

  \authorblockA{
    \authorrefmark{1}
    Institute of Information Science, Academia Sinica, Taipei, Taiwan \\
    E-mail: \{jeremycchsu, hwanght, tedwu, whm\}@iis.sinica.edu.tw
  }

  \authorblockA{
    \authorrefmark{2}
    Research Center for Information Technology Innovation, Academia Sinica, Taipei, Taiwan \\
    E-mail: yu.tsao@citi.sinica.edu.tw
  }
}

\maketitle
\thispagestyle{empty}

\begin{abstract}
  We propose a flexible framework for spectral conversion (SC) that facilitates training with unaligned corpora.
  Many SC frameworks require parallel corpora, phonetic alignments, or explicit frame-wise correspondence for learning conversion functions or for synthesizing a target spectrum with the aid of alignments. 
  However, these requirements gravely limit the scope of practical applications of SC due to scarcity or even unavailability of parallel corpora. 
  We propose an SC framework based on variational auto-encoder which enables us to exploit non-parallel corpora. 
  The framework comprises an encoder that learns speaker-independent phonetic representations and a decoder that learns to reconstruct the designated speaker. 
  It removes the requirement of parallel corpora or phonetic alignments to train a spectral conversion system.
  We report objective and subjective evaluations to validate our proposed method and compare it to SC methods that have access to aligned corpora.
\end{abstract}

\section{Introduction}
  Voice conversion is a technique that converts the perceived identity of speaker of a given utterance.
  A typical case is that, when one wants to convert his or her voices into a celebrity's, it is required that linguistic contents and other speaker-unrelated information remain unchanged after conversion.
  A complete voice conversion system involves many tasks. In this study, we devote our focus on spectral conversion (SC) and leave inspection on prosody outside the scope of this paper.

  A wide variety of techniques have been applied to spectral conversion, including Gaussian mixture models (GMM)
  \cite{Toda07, GMMTakamichi, Hwang13-GMM-VC},
  frequency warping
  \cite{Erro10-FreqWarp, Godoy12-FreqWarp},
  deep neural networks (DNN)
  \cite{Desai10-NN-VC, Nakashika13-NN-VC, Hwang15-NN-VC},
  and exemplar-based approaches
  \cite{Wu14-NMF-VC, Wu16-LLE-VC}.
  Most of these methods demand aligned source-target pairs of frames or alignments of phonetic states to train conversion functions or adaptation transformations.
  The most widely adopted approach is to align source and target frames using dynamic time warping (DTW) technique.
  However, DTW fails to work if parallel corpora are unavailable.

  Many techniques have been conceived to align source and target frames in non-parallel corpora.
  The most intuitive way is to apply a speech recognizer to the utterances, and proceed with explicit alignment or model adaptation
  \cite{DBLP:conf/apsipa/DongYLEHMTLL15, DBLP:conf/icassp/ZhangTTW08}.
  Applying speech recognizers to each utterance gives every frame a phonetic label (usually of phonetic states).
  It is particularly suitable for model-based voice conversion techniques because they can readily utilize these labeled frames
  \cite{DBLP:conf/icassp/SongZZ13}.
  The problem with this frame-wise, model-based approach is that it does not apply to cross-lingual conditions, which require a more general form of alignment.
  To this end, the INCA algorithm and related methods
  \cite{DBLP:journals/taslp/ErroMB10a, DBLP:conf/icassp/BenistyMC14,DBLP:conf/icassp/Agiomyrgiannakis16}
  were proposed to iteratively seek frame-wise correspondence using converted surrogate frames.
  Another attempt is to separately build frame clusters for the source and the target, and then set up a mapping between them
  \cite{DBLP:conf/interspeech/NeySBH04}.

  Let us ponder upon the roles these alignment techniques play in the task of voice conversion.
  Consider a speech corpus of a source speaker $s$ and a target $t$. 
  The subset $\Xs = \{\xsn\}_{n=1}^{N_s}$ represents all the frames from the source
  and $\Xt = \{\xtn\}_{m=1}^{N_t}$ are those from the target, 
  where $N_s$ and $N_t$ are the total number of frames of the source and the target, respectively. 
  
  The first and probably the most general kind of alignment is frame-wise.    
  It seeks index pairs $(n, m)$ such that $\xsn$ and $\xtn$ have similar phonetic contents.
  For simplicity, we assume that the correspondence is a function
  (though it is not in most cases).
  Frame-wise alignment is therefore a function of $\xsn$ that yields a corresponding $\xtn$.
  Under this circumstance, alignment is not only necessary, but also nearly sufficient for SC because SC also pursues similar functions.
  
  The second kind is frame-to-model alignment attained with the help of speech recognizers.
  It assumes that every frame corresponds to a (phonetic) model (or, equivalently, a cluster).
  The alignment is thus $(n, k)$ pairs where $k$ is the model index of a phonetic state.
  Conversion is then the transformation function that inputs a model from the source, and outputs a model from the target.
  For the purpose of conversion, alignment is also necessary under this scenario.

  In contrast, the factor of speaker plays a rather implicit role in voice conversion.
  For example, in most pair-wise SC (one source and one target), speaker identity is only responsible for designating a frame to the input (if it is from the source) or to the output (if otherwise).
  It is curious that we build voice conversion systems without explicitly exploiting speaker-dependent factors considering the purpose.

  We propose a framework that directly exploits speaker identity to build SC systems without explicitly aligning source and target frames.
  Our proposed formulation decomposes conversion into encoding and decoding stages, and renders conversion a controlled version of self-reconstruction.
  With this self-reconstruction formulation, aligned frame pairs or even parallel corpora are no longer necessary for SC tasks.
  Our experiments showed that its performance is comparable to baseline systems, substantiating this nascent framework for general SC tasks.

  The rest of this paper is organized as follows.
  In Sec. \ref{sec:ours}, we describe the inspiration and the concepts, and elaborate our methods.
  Experimental settings and results are collected in Sec. \ref{sec:exp} to validate our proposed framework.
  Finally, we conclude our paper in Sec. \ref{sec:conclusion}.

\section{The Proposed Method}
\label{sec:ours}
  The proposed method is inspired from an analogous work on generating hand-written digits
  \cite{DBLP:journals/corr/KingmaRMW14, Kingma14-VAE}.
  The authors of \cite{DBLP:journals/corr/KingmaRMW14} attempted to extract writing style and digit identity from an image of handwriting and to re-synthesize the image with the extractive.
  Basically what the framework offers is an explanatory model of an observed variable and two causal latent factors: identity and variation.
  We hypothesize that the explanatory model behind speech frames coincides with that of hand-writing images.
  For a hand-written digit, 
  the identity is the nominal number and the variation is the hand-writing style.
  For a speech frame, 
  the identity could be the speaking source and the variation could be the phonetic content.

  \subsection{Auto-encoder Reformulation for SC from Unaligned Data}
    Given spectral frames $\{\xsn\}_{n=1}^{N_s}$ from the source speaker and 
    those $\{\xtn\}_{m=1}^{N_t}$ from the target, 
    conventional SC seeks to estimate conversion functions such that
    \begin{equation}
    \label{eq:traditional-VC}
      \xtnhat = f(\xsn), 
    \end{equation}
    where $\f$ is a conversion function.
    In most SC systems, 
    speaker identity (subscripts $s$ and $t$) are treated implicitly; 
    for example, in \eq{eq:traditional-VC}, the source is always the input while the target is always the desired output.

    We explicitly incorporate a speaker representation $\yn$ into the SC formulation.
    Firstly, the conversion function is reformulated as an auto-encoder.
    The encoder $\fp(\cdot)$ is designed to be speaker-independent; 
    it ignores speaker identity of an incoming frame 
    (so $\xsn$ and $\xtn$ can now be expressed by $\xn$), 
    and converts an observed frame into speaker-independent latent variable:
    \begin{equation}
    \label{eq:z-eq-qx}
      \zn = \fp(\xn),
    \end{equation}
    where $\zn$ is a latent variable (or \emph{code} in auto-encoder terminology).
    Presumably, $\zn$ contains information that is irrelevant to speaker, such as phonetic variations.
    We refer to $\zn$ as phonetic representation later in this paper for convenience (though $\zn$ might cover more than phonetic traits).

    Next, we need a decoder $\ft(\cdot)$ to reconstruct speaker-dependent frames.
    For that purpose, we introduce the speaker representation $\yn$ as another latent variable, and concatenate it to $\zn$.
    The decoder then utilizes the joint vector $(\yn, \zn)$ to reconstruct a speaker-dependent frame $\xhat$  ($\xsnhat$ or $\xtnhat$, depending on $\yn$):
    \begin{equation}
    \label{eq:x-eq-p-zy}
      \xhat = \ft(\zn, \yn).
    \end{equation}

\begin{figure}[t]
      \includegraphics[width=0.25\textwidth]{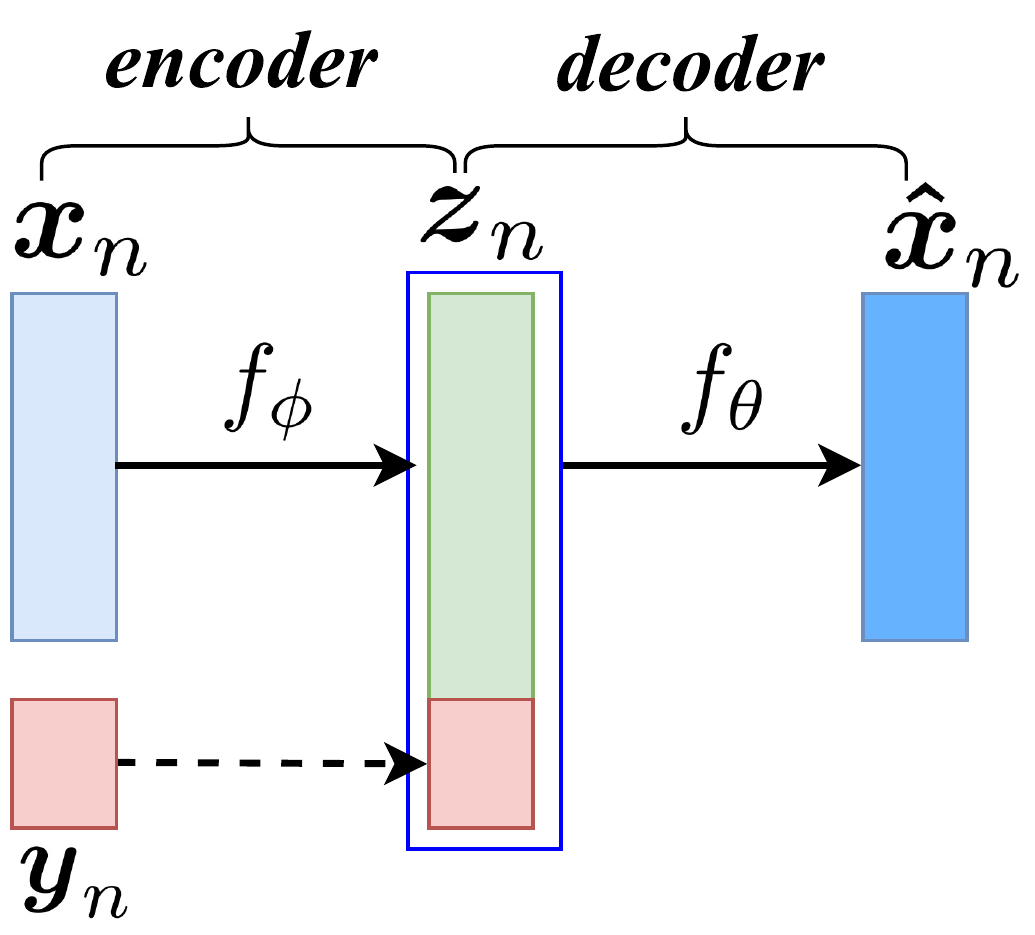}
      \centering
      \caption{Illustration of VAE-based non-parallel spectral conversion. The dashed line means copying. The latent variable $\zhat$ and $\yn$ are concatenated.}
      \label{fig:arch}  
    \end{figure}

    To sum up, reformulation is achieved by substituting $\f$ in \eq{eq:traditional-VC} with $\ft(\cdot)$ in \eq{eq:x-eq-p-zy} and then $\zn$ with \eq{eq:z-eq-qx}:
    \begin{equation}
      \xhat = \hat{f}(\xn, \yn) = \ft(\zn, \yn) = \ft(\fp(\xn), \yn).
    \end{equation}
    Alignment plays no roles in this formulation because the encoder-decoder pair accepts a frame $\xn$ and a speaker representation $\yn$ on a frame-wise basis.
    It then puts together the phonetic representation $\zn$ and the speaker representation $\yn$ to synthesize a frame $\xnhat$.
    Fig. \ref{fig:arch} depicts the structure.

    The framework's viability relies on two assumptions.
    First, we assume that speaker representation and phonetic representation can be decoupled from a given frame.
    Second, we assume that the decoder can blend the two factors (phonetic and speaker identity) to synthesize a spectral frame.

    \subsection{Architecture}
      We modify variational auto-encoder (VAE)
      \cite{Kingma14-VAE, DBLP:journals/corr/KingmaRMW14}
      to tackle the problem of SC from unaligned data.
      A VAE is a directed probabilistic model realized in the form of neural networks.
      We choose the variational over vanilla auto-encoder because the former has a more understandable model for the latent space and better a regularization property.
      We have described the basic concepts and the auto-encoder part in the previous section.
      Now we shall elaborate some of the details, including the training objective and the inference process.

      We regard SC as a generative process of VAE, and therefore try to maximize joint log-probability of individual frames:
      \begin{equation}
      \label{eq:logp}
        \log \p(\X) = \summation_{n=1}^{N} \log \p(\xn).
      \end{equation}
      The individual log-probability of VAE can be re-written as:
      \begin{equation}
      \label{eq:logp-kld-lb}
      \begin{split}
        \log \p(\xn) & = \KLD{\q(\zn|\xn)}{\pTrue(\zn|\xn)} \\
                     & + \LB{\xn},
      \end{split}
      \end{equation}
      where $\q(\cdot)$ is the variational posterior and $\pTrue(\cdot)$ is the true posterior.
      The first right-hand-side (RHS) term $\KLD{\cdot}{\cdot}$ is the Kullback-Leibler divergence (KLD) of the approximate from the true posterior.
      The second RHS term is called the variational lower bound on the marginal probability and can be further rewritten as:
      \begin{equation}
      \label{eq:lb}
      \begin{split}
        \LB{\xn} = - & \KLD{\q(\zn|\xn)}{\pTrue(\zn)} \\
                   + & \E{\q(\zn|\xn)}{\log \p(\xn|\zn)}.
      \end{split}
      \end{equation}
      Direct optimization of \eq{eq:logp-kld-lb} is usually intractable, so we instead take the variational lower bound \eq{eq:lb} as our objective function.
      The goal is to differentiate and optimize the lower bound w.r.t. the encoder parameters $\phis$ and decoder parameters $\thetas$.
      We will first descibe how to estimate the expectation term which is the cost of induced by latent space modeling.
      Then, we will derivate the closed-form expression for the KLD term.

      \subsubsection{Estimating the Expectation Term}
      Sampling methods are frequently adopted to estimate the expectation term in \eq{eq:lb}:
      \begin{equation}
                \E{\q(\zn|\xn)}{\log \p(\xn|\zn, \yn)} 
        \approx \summation_{l=1}^{L} \log \p(\xn|\zn, \yn),
      \end{equation}
      where $L$ is the number of samples drawn per frame.
      However, naive sampling is usually problematic, so we resort to the re-parameterization trick
      \cite{Kingma14-VAE}.
      We sample from the distribution of $\zn$ by generating a standard normal random variable and apply a data-driven deterministic function to it:
      \begin{equation}
      \begin{split}
      \label{eq:sampling}
        \zhat &\sim \N(\zn; \zmu, diag(\zsig^2)), \\
        \eps &\sim \N(\0, \I), \\
        \zmu &= \fpOne(\xn) = \zn, \\
        \log{\zsig} &= \fpTwo(\xn), \\
        \Rightarrow ~ \zhat &= \fp(\xn) = \zmu + \eps \circ \zsig,
      \end{split}
      \end{equation}
      where $\circ$ denotes Hadamard (element-wise) product,
      $\fpOne$ and $\fpTwo$ are non-linear functions made of feed-forward neural networks,
      and $\phis = \{\phiOne, \phiTwo\}$ is the set of encoder parameters.
      With re-parameterization, the expectation term in \eq{eq:lb} is approximated by:
      \begin{equation}
      \begin{split}
      \label{eq:reparam}
          &~ \E{\q(\zn|\xn)}{\log \p(\xn|\zn, \yn)} \\
        = &~ \E{\N(\zn; \zmu, diag(\zsig)}{\log \p(\xn|\zn, \yn)} \\
        = &~ \E{\N(\eps; \0, \I)}{\log \p(\xn|\zhat, \yn)} \\
        \approx & \summation_{l=1}^{L} \log \p(\xn|\zhat, \yn).
      \end{split}
      \end{equation}

      We simplify \eq{eq:reparam} by setting $L$ to 1, resulting in the final approximated objective function of an individual frame:
      \begin{equation}
      \begin{split}
      \label{eq:obj-sub}
      \LBhat{\xn} = & - \KLD{\q(\zhat|\xn)}{\pTrue(\zn)} \\
                    & + \log \p(\xn|\zhat, \yn). 
      \end{split}   
      \end{equation}

      \subsubsection{Modeling the Latent Space}
      The prior distribution of latent variable $\zn$ can be thought of as our imagination of the origin of the visible variable $\xn$,
      and the KLD in \eq{eq:lb} can be deemed as a term that regularizes the latent variable not to distribute too differently from the chosen prior of $\zn$.
      Our choice of $\zn$ is an isotropic standard normal distribution, which concords with
      \cite{Kingma14-VAE}.    
      Thanks to the choice of Gaussian latent variable, the KLD term (cost of the latent variable) can be evaluated in closed-form:
      \begin{equation}
      \begin{split}
      \label{eq:kld-term}
        & -D_{KL}(\q(\zn|\xn) || \pTrue(\zn)) \\
      = & -D_{KL}(\N(\zhat; \zmu, diag(\zsig)) || \N(\zn; \0, \I)) \\
      = & \half \summation_{d=1}^{D} (1 + \log{\zvard} -\zmud^2 -\zvard),
      \end{split}  
      \end{equation}
      where $D$ is the dimension of the latent space.

      \subsubsection{Modeling the Visible Space}
      We assume that the visible variable of our features (log-spectrum) obeys Gaussian distribution with a diagonal variance matrix:
      \begin{equation}
      \label{eq:vis-gauss}
      \begin{split}
        \xhat & \sim \N(\xn; \xmu, \xsig), \\
        \xmu & = \ftOne(\zn, \yn), \\
        \log{\xsig} & = \ftTwo(\zn, \yn),
      \end{split}   
      \end{equation}
      where $\ftOne$ and $\ftTwo$ are non-linear functions made of feed-forward neural networks,
      and $\thetas = \{\thetaOne, \thetaTwo\}$ is the set of decoder parameters.
      The log-probability term in \eq{eq:obj-sub} can therefore be expressed in closed-form:
      \begin{equation}
      \label{eq:log-prob-term}
      \begin{split}
          & \log \p(\xn|\zhat, \yn) = \log \N(\xn; \xmu, \xsigma) \\
        = & -\half \summation_{d=1}^{D} \Big( 
             \log(2 \pi \xvard) 
           + \dfrac{(x_d - \xmud)^2}{\xvard} \Big),
      \end{split}   
      \end{equation}
      where $D$ is the dimension of the visible (feature) space.
      
      The final objective function can be obtained by substituting 
      \eq{eq:log-prob-term} and \eq{eq:kld-term} into \eq{eq:obj-sub}.
      Training is equivalent to iteratively finding the parameters that maximize the variational lower bound:
      \begin{equation}
        \{\thetas^*, \phis^*\} = \argmax_{\thetas, \phis}~ \LBhat{\X}.
      \end{equation}
      We use stochastic gradient descent (SGD) for optimization in our implementation.

    \subsubsection{Conducting Conversion}
    Spectral conversion is straightforward since we merely have to specify $\yn$ that corresponds to the desired target. 
    The encoder first transforms the input frame into a latent representation, 
    and next, the decoder transforms ($\zn$, $\yn$) into $\xhat$. 
    Note that sampling is not needed in the conversion phase.

  \subsection{Training Procedures}
    The training procedures of a VAE-based SC system differ from those of a conventional system.
    First, a speaker representation $\yn$ has to be introduced to train the decoder.
    It can be as simple as a one-hot vector, pre-defined for each speaker,
    or a probability vector.
    We will describe the speaker representation using the one-hot vector in the following paragraphs.
    Second, training a VAE involves \emph{sampling} from a probability distribution of a latent variable $\zn$, and this means injection of stochasticity.
    Third, training the VAE is \emph{point-wise} as opposed to \emph{pair-wise} in conventional systems. 
    That is, $\xsn$ and $\xtn$ are no longer discriminated with the former being input and the latter being output; they are both viewed as $\xn$.
    The source and the target sets are deemed as one unified set, and the speaker identity of each frame is added to the training set:
    \begin{equation}
    (\mat{X}, \mat{Y}) = \{(\xn, \yn)\}_{n=1}^{N = N_s + N_t}.
    \end{equation}

    Since our proposed method is an auto-encoder that reconstructs the input, training is conducted by feeding a pair of a spectral frame and its corresponding speaker identity $(\xn, \yn)$ into the auto-encoder. 
    Note that the speaker identity of every utterance is always known in most speech corpora for the purpose of voice conversion.
    Hence, the speaker identity of every frame is also known.
    Consequently, our proposed framework can explicitly utilize speaker identity as an additional input.

    The encoder treats every incoming frame in the same way as if the speaker identity is unknown;
    It transforms an input frame into a speaker-independent latent phonetic representation.
    As the encoder receives frames from both the source and the target, it  cultivates the ability of speaker-independent encoding.

    Subsequently, the decoder reconstructs the input from the latent representations.
    It first samples from the distribution of the code (latent variable $\zn$), and then reconstructs the input with the aid of speaker representation $\yn$.

    Finally, costs defined on the visible and the latent variables are computed and jointly optimized, 
    and the network parameters are updated iteratively.
    The training procedures would terminate when it reached maximum generation probability.

\section{Experiments}
\label{sec:exp}
  \subsection{Experimental Settings}
    \subsubsection{The VCC2016 Speech Corpus}
      The proposed SC system was evaluated on a parallel English corpus from the Voice Conversion Challenge 2016 \cite{VCC2016}. 
      There are 5 male and 5 female speakers in this corpus. 
      Each speaker has 150 utterances as the training set and 12 utterances as the evaluation set. 
      The evaluation set was aligned, and the objective evaluations were conducted on this set.
      Five out of the ten speakers are designated to be the conversion targets (2 female and 3 male speakers) and the other five sources (3 female and 2 male speakers). 
      The testing set comprises 54 utterances per target speaker, and we use this testing set to generate converted voices for subjective evaluation.

      We conducted experiments on a subset of the speakers. 
      Two speakers were chosen as sources (SF1 and SM1) and another two as targets (TF2 and TM3).
      We further divided the training set into disjoint (non-parallel) subsets to train one of the VAE variants in Sec. \ref{sec:variants}.
      We reported two types of spectral conversion: intra-gender and cross-gender.

    \subsubsection{Feature Sets}

      We used the STRAIGHT toolkit \cite{Kawahara99-STRAIGHT} to exctract  speech parameters, including the STRAIGHT spectra (SP for short), aperiodicity (AP), and pitch contours (F0). 
      The FFT length was set to 1024, so the resulting AP and SP were both 513-dimensional. 
      The frame shift was 5 ms and the frame length was 25 ms. 
      We did not incorporate contextual or dynamic features into the feature set. 
      Every input frame of SP was normalized to unit-sum, and the normalizing factor (energy) was taken out as an independent feature and was not modified.
      The SP was converted using our proposed method or the baseline systems. 
      Note that we further applied logarithm on SP in our proposed method, whereas we used linear (non-negative) SP in the baseline systems.
      All systems converted F0 using the same linear mean-variance transformations on log-F0 domain. 
      The AP was kept unmodified.
      After spectral conversion, energy was compensated back to SP, and STRAIGHT took in all the parameters to synthesize utterances.


  \subsection{Baseline Systems}
    The baseline systems were built on Exemplar-based Non-negative Matrix Factorizations (ENMF) using parallel data. The systems were similar to those described in \cite{Wu14-NMF-VC}. The dictionaries were 512 or 3000 randomly selected source-target pair frames and thus the baseline systems were labeled ENMF-512 and ENMF-3000, respectively.

    In baseline systems, each parallel training set was aligned using dynamic time warping (DTW) with 24-ordered Mel-cepstral coefficients (MCC) extracted from SP.
    After alignment, the length of a source utterance remained the same while some frames from the target were duplicated or decimated.
    Next, energy-based voice activity detection (VAD) was used to exclude the silence segments.

    These baseline systems require no training. 
    They convert a spectral frame by optimizing self-reconstruction criterion; in the process, they obtain an activation matrix which is the weights of linear combination of dictionary bases.
    The activation matrix is then applied to a parallel dictionary of the target to convert into his or her spectral frame.
    As a result, conversion can be conducted on-line.

  \subsection{Variational Auto-Encoders}
    \subsubsection{Configurations and Hyper-parameters}
      The encoder and the decoder were feed-forward neural networks with 2 hidden layers, each with 512 nodes.\footnote{We implemented our systems using Tensorflow's Python API
      \cite{tensorflow2015-whitepaper}.}
      Rectifier linear units (ReLU) \cite{Nair10-ReLU} were applied to each layer to provide non-linearity (except for output layers $\zn$ and $\xhat$, which were linear).
      The latent (phonetic) space was 64-dimensional.
      The size of a mini-batch was 128.
      The optimizer was ADAM \cite{Kingma15-Adam}.
      The dimension of speaker representation is identical to the number of speakers in the training subset (2 for one-to-one conversion and 4 for a unified, multiple-speaker conversion).

      The visible space of log-spectrum feature was modeled by a Gaussian distribution (as in \eq{eq:vis-gauss}).
      We ignored variance modeling and adopted an identity matrix for it because variance did not affect the generative process in our system.
      The desired prior distribution for the latent variables was an isotropic standard normal distribution (as in \eq{eq:sampling}).
      
    \subsubsection{Three Variants}
    \label{sec:variants}
      We report SC results of three variants of the proposed framework.
      The first system, referred to as VAE-pair, was built from a single source and a single target, each with 150 utterances.
      The second, labeled VAE-multi, was built from the whole training subset of 4 speakers.
      The last, labeled VAE-disj, was built from non-parallel data.
      Its training set consisted of the first 75 utterances of the source and the other 75 from the target.
      We shall clarify three things.
      First, VAE-pair and VAE-multi were trained using \emph{parallel but unaligned data} while VAE-disj was trained using \emph{non-parallel data}.
      Second, the size of the training sets of VAE-disj was roughly halved because the set of sentences from the source and that from the target were mutually exclusive.

  \subsection{Objective Evaluations}
\begin{figure}[t]
      \includegraphics[width=0.5\textwidth]{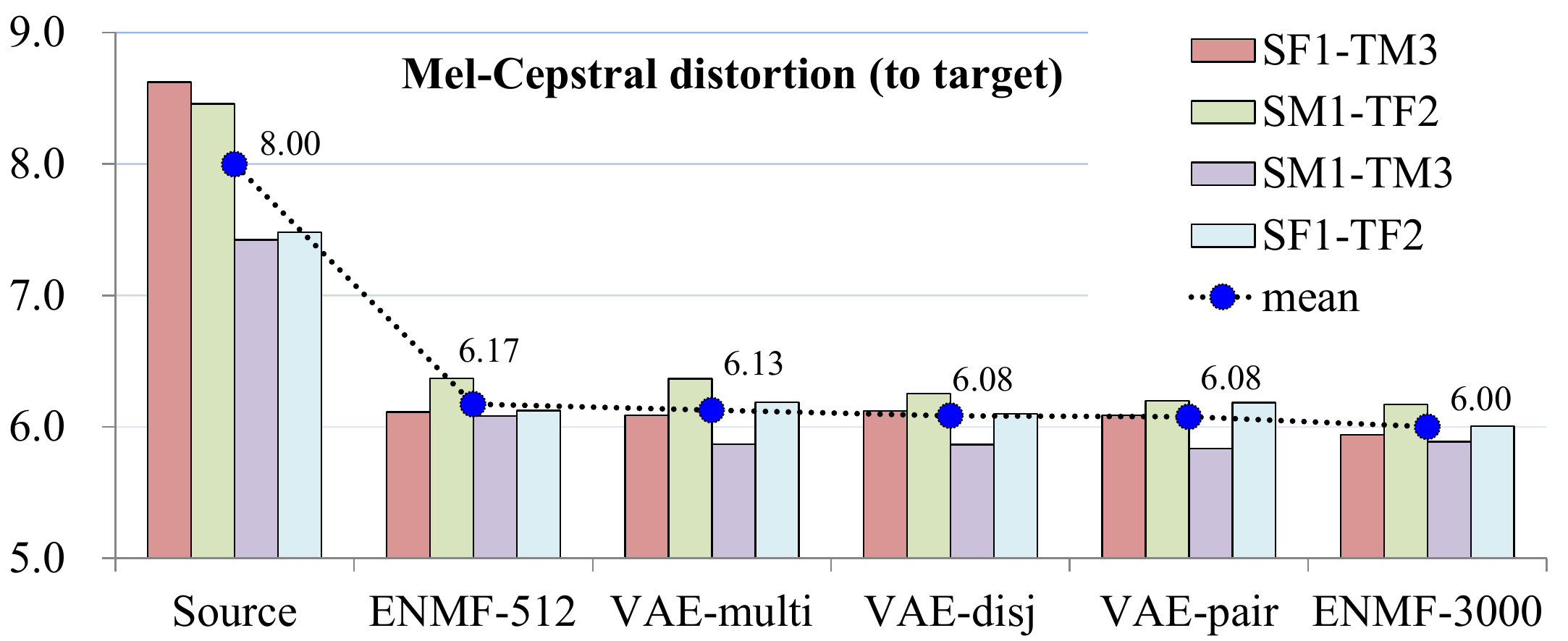}
      \centering
      \caption{Mean Mel-cepstral distortion of the proposed method compared to baseline systems that have access to alignment information. The figure is arranged according to mean MCD.}
    \label{fig:mcd}
    \end{figure}

    We visualize mean Mel-cepstral distortion (MCD) values on the evaluation set in Fig. \ref{fig:mcd}. 
    Our proposed methods trained on unaligned data performed on par with the baselines which utilized aligned frames.
    The results might imply that all the systems achieved comparable level of performance.
    As MCD was not a representative indicator for perception, we further conducted subjective evaluations on voice quality and similarity.

  \subsection{Subjective Evaluations}
\begin{figure}
    \centering
      \begin{subfigure}[b]{0.49\textwidth}
       \includegraphics[width=1\linewidth]{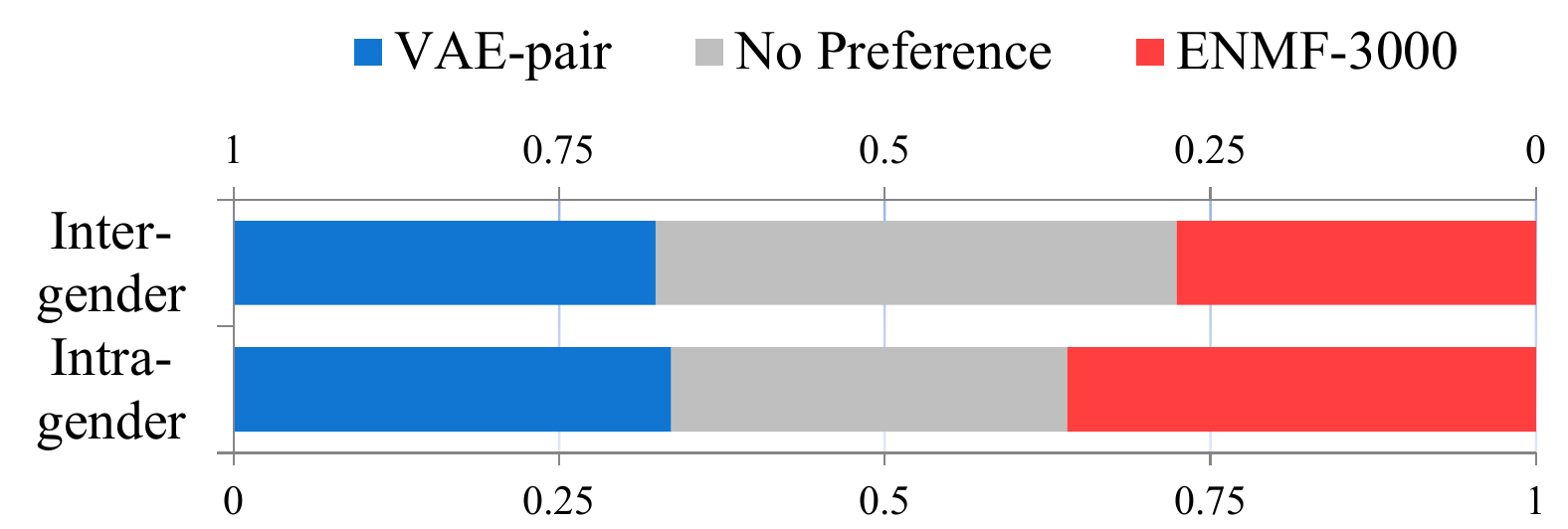}
       \caption{Preference on voice quality.}
       \label{fig:abx-qua} 
      \end{subfigure}
      \begin{subfigure}[b]{0.49\textwidth}
       \includegraphics[width=1\linewidth]{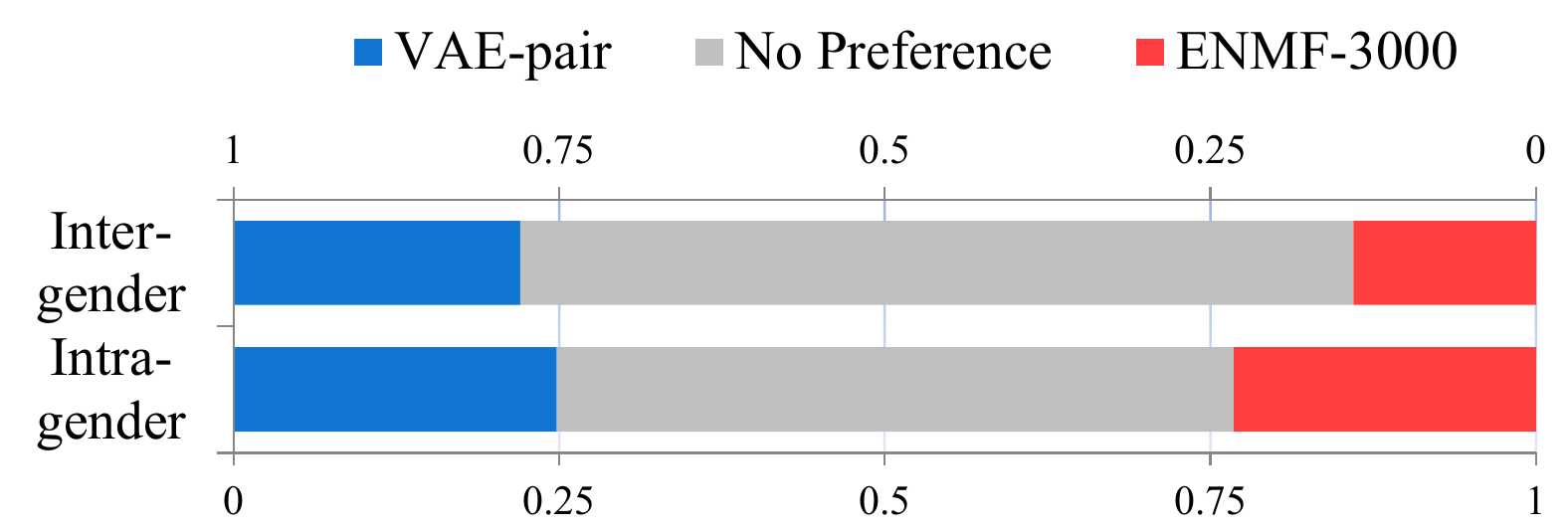}
       \caption{Preference on similarity.}
       \label{fig:abx-sim}
      \end{subfigure}
      \caption{Preference on voice quality and similarity to the target. The target is TF2 (female) and the source is SF1 and SM1 for intra- and inter-gender conversion, respectively.}
      \label{fig:abx}
    \end{figure}

    As for subjective evaluation, we chose ENMF-3000 as the baseline because it offered higher quality of synthetic voice than ENMF-512.
    We evaluated our proposed method (VAE-pair) by listening tests. 
    Ten listeners were invited to evaluate the results.
    We divided our experiments into inter- and intra-gender conversion.
    Every listener was asked to evaluate a mean opinion score (MOS) on voice quality and ABX tests on voice quality and target similarity.
    The results are shown in
    Fig. \ref{fig:abx}.

    The ABX test on target similarity revealed that both systems performed at a comparable level.
    This result was anticipated, and was consistent with the MCD objective evaluation.
    As for voice quality, our proposed method also achieved similar level as the ENMF-3000 baseline.
    VAE-pair achieved 2.76 MOS (with standard deviation 0.44) while ENMF-3000 achieved 2.75 MOS (with standard deviation 0.50).
    This result was rather encouraging since we initially conjectured that the performance degradation would be somewhat higher because VAE-pair used unaligned training data.
    Note that the voice quality of ENMF-3000 was rather acceptable (unlike that of ENMF-512, which was at the brink of satisfaction). More subjective evaluations on VAE-multi and VAE-disj will be conducted in our future work.

  \subsection{Training from Non-parallel Corpora}
     We were surprised that the performance of VAE-disj was  around the same as VAE-pair in the objective evaluations (cf. Fig. \ref{fig:mcd}),
    since the training condition of the former was apparently harsher than the latter.
    While the experiment verified that our framework was applicable to non-parallel corpora,
    it also pointed out some issues.
    For example, the capability of the models might not have been fully exploited because the size of the training set of VAE-pair was twice that of VAE-disj.
    We shall investigate the cause more profoundly in the future.

  \subsection{Toward Many-to-Many Voice Conversion}
  \label{sec:m2m}
    From Fig. \ref{fig:mcd}, we also observed that the performance of VAE-multi was close to that of VAE-pair in the objective evaluations.
    It is interesting in two aspects.
    First, the two systems share nearly identical setting of hyper-parameters.
    The model of VAE-multi had to learn much more complex functions as it consolidated many pair-wise systems into one.
    Second, VAE-multi is virtually able to convert any of the 12 permutations of the 4 speakers, i.e., VAE-multi consolidates 12 systems into one.
    Its ability is evocative of many-to-many (M2M) voice conversion.

    We conjectured that we could be only one step behind M2M conversion.
    An M2M conversion system has two requirements.
    First, it must be capable of convert an arbitrary, even unseen, source to a given target.
    Second, it must be able to convert a source to a target that never appears in the training phase, but has limited resources during conversion.
    Conceptually, our framework has the ability to accommodate M2M tasks.
    This could be achieved by introducing a speaker recognition network (in the form of another encoder) to replace the given speaker representation (one-hot vector in our case).
    Or, the speaker representation could be in other forms.
    Once the specker representation of the unknown target speaker is obtained from the limited speech, it is likely that the decoder can blend speaker and phonetic representations to synthesize a speaker-dependent spectral frame, thus achieving M2M conversion.

\section{Conclusions}
\label{sec:conclusion}
  In this paper, we have introduced a VAE-based SC framework that is able to utilize
  unaligned data.
  It was an attempt toward training without the need of explicit alignment.
  Objective and subjective evaluations validated its ability to convert spectra, and the performance of the proposed method is comparable to baseline systems that have access to aligned data.
  We will continue to improve its performance, 
  investigate its ability to accommodate many-to-many voice conversion, 
  and generalize it to more tasks.

\bibliographystyle{IEEEtran}
\bibliography{jrm}

\end{document}